\documentclass{Interspeech2024}

\interspeechcameraready

\usepackage{inputenc, amsmath, amssymb, xfrac, fontenc, bbm, graphicx, amsfonts, cite, hyperref, multirow, multicol, booktabs, algorithm, algpseudocode}
\usepackage{subfigure}
\usepackage{adjustbox}
\def\mathbi#1{\textbf{\em #1}}

\title{Query-by-Example Keyword Spotting Using Spectral-Temporal Graph Attentive Pooling and Multi-Task Learning}

\name[affiliation={}]{Zhenyu}{Wang}
\name[affiliation={}]{Shuyu}{Kong}
\name[affiliation={}]{Li}{Wan}
\name[affiliation={}]{Biqiao}{Zhang}
\name[affiliation={}]{Yiteng}{Huang}
\name[affiliation={}]{Mumin}{Jin}
\name[affiliation={}]{Ming}{Sun}
\name[affiliation={}]{Xin}{Lei}
\name[affiliation={}]{Zhaojun}{Yang}
\address{
  Meta AI}
\email{zhenyu.wang@utdallas.edu, zhaojuny@meta.com}
\keywords{Query-by-example Keyword Spotting, Conformer, LicoNet, Spectral-temporal Attentive Pooling, Additive Angular Margin, SoftTriplet}

\begin{document}

\maketitle
 
\begin{abstract}
Existing keyword spotting (KWS) systems primarily rely on predefined keyword phrases. However, the ability to recognize customized keywords is crucial for tailoring interactions with intelligent devices. In this paper, we present a novel Query-by-Example (QbyE) KWS system that employs spectral-temporal graph attentive pooling and multi-task learning. This framework aims to effectively learn speaker-invariant and linguistic-informative embeddings for QbyE KWS tasks. Within this framework, we investigate three distinct network architectures for encoder modeling: LiCoNet, Conformer and ECAPA\_TDNN. The experimental results on a substantial internal dataset of $629$ speakers have demonstrated the effectiveness of the proposed QbyE framework in maximizing the potential of simpler models such as LiCoNet. Particularly, LiCoNet, which is 13x
more efficient, achieves comparable performance to the computationally intensive Conformer model ($1.98\%$ vs.~$1.63\%$ FRR at $0.3$ FAs/Hr).
\end{abstract}

%\noindent\textbf{Index Terms}: Query-by-example Keyword Spotting, Conformer, LicoNet, Spectral-temporal Attentive Pooling, Additive Angular Margin, SoftTriplet

\section{Introduction}
\label{sec:intro}
A keyword spotting (KWS) system serves the purpose of detecting a predetermined keyword within a continuous real-time audio stream. This capability is pivotal in facilitating interactions between users and voice assistants. The introduction of a customized KWS system, which empowers users to define their own keywords, offers a substantial degree of flexibility and personalization in user experiences. However, in addition to the typical difficulties associated with KWS, such as maintaining a small memory footprint and minimizing latency, the customization process also introduces a significant challenge: user-defined keyword phrases may not align with the distribution of the training data, resulting in inferior detection performance. One approach for customized KWS involves the use of Query-by-Example (QbyE) techniques \cite{chen2015query}. In this context, the KWS system utilizes audio samples of keywords provided by users to generate fixed-length embeddings. These embeddings are then employed to assess the similarity between test samples and enrolled keywords within the embedding space, ultimately determining the presence of a keyword. 

Extensive research has explored the application of various neural network architectures in the context of pre-defined KWS tasks. In \cite{chen2018compact} \cite{guo2018time}, an encoder-decoder model structure has been proposed, where an acoustic encoder generates senone-level posteriors, and a decoder is independently trained to interpret the encoder outputs as keyword phrases. Following the trend of end-to-end modeling, researchers have also suggested building an end-to-end KWS system that directly outputs detection without distinguishing the encoder and decoder in the training \cite{shan2018attention} \cite{wang2019adversarial}. Despite achieving high detection performance, training a dedicated model for predefined keywords typically requires substantial target data to ensure effectiveness.

%\cite{shan2018attention,wang2019adversarial} employed recurrent neural networks (RNNs) with attention layers to build an end-to-end KWS system. %Our previous work proposed to generate adversarial examples for data augmentation to enable better model generalization in data-sparse scenarios \cite{10094750}.

Customized KWS systems alleviate data requirements and offer flexibility by supporting keywords beyond a predefined set.
Early endeavors on QbyE KWS tasks rely on a pre-trained Automated Speech Recognition (ASR) system. Specifically, the acoustic model of an ASR first generates 
phonetic posteriors for an audio stream, and dynamic time warping (DTW) is employed to measure the similarity between the posterior sequences of enrolled keywords and testing samples \cite{hazen2009query} \cite{zhang2009unsupervised} \cite{anguera2013memory}. The CTC-based KWS approach extends this framework 
by employing CTC forwarding to evaluate the likelihood of each keyword hypothesis given an audio sample. It then generates a final detection score by aggregating the scores from the N-best list \cite{bluche2020small} \cite{zhuang2016unrestricted}.
%that scores each hypothesis of the keyword given an audio sample through CTC forwarding and generates a final detection score by aggregating the scores of the $N$-best list \cite{bluche2020small} \cite{zhuang2016unrestricted}.  

Chen et.~al have proposed LSTM-based encoder modeling for learning acoustic embeddings of customized keywords and have shown promising results \cite{chen2015query}. 
%\cite{lugosch2018donut} \cite{bluche2020small} \cite{zhuang2016unrestricted} use RNNs with Connectionist Temporal Classification (CTC) loss to detect the keyword by aggregating scores of all hypotheses given an input audio sample.
\cite{huang2021query} further improves encoder modeling by incorporating multi-head attention for feature extraction, using a normalized multi-head attention module for feature aggregation, and integrating SoftTriplet loss to enhance discrimination capabilities, resulting in improved performance and capabilities.
However, attention-based models are accompanied by significant computational demands, posing a high runtime memory burden when deployed on hardware. This characteristic makes them unsuitable for an always-on KWS system. To address this challenge, Huang et.~al have made an initial attempt by replacing the attention mechanism with the MLPMixer architecture in the QbyE KWS tasks \cite{huang2022qbye}. Despite the improved performance and efficiency compared to ViT, the frequent utilization of matrix transpose operations can still cause significant computation overhead on hardware. 

In this study, we introduce an effective framework for QbyE KWS. The framework employs spectro-temporal graph attentive pooling (GAP) \cite{tak2021end} and multi-task learning to facilitate informative embedding learning. GAP demonstrates strong capability in comprehending complex relationships within spectral-temporal data. The multi-task learning is designed to minimize the loss of classifying words and phonemes but to maximize the loss of distinguishing speakers, aiming to enhance the distinctiveness of words and phonemes while simultaneously reducing the speaker variability in learned embeddings. The loss function incorporates Additive Angular Margin (AAM) loss and SoftTriplet loss, both widely employed in tasks such as speaker recognition \cite{xiang2019margin} and face recognition \cite{deng2019arcface}. 
%Through this framework, we aim to advance customized KWS for improved performance. 
Within this framework, we investigate three distinct network architectures for encoder modeling: LiCoNet, Conformer and ECAPA\_TDNN.  
Our experimental results, conducted on a substantial internal dataset of $629$ speakers, showcase  the effectiveness of the proposed QbyE framework in maximizing the potential of simpler models. Notably, within this framework, LiCoNet, which is $13$x more efficient, achieves comparable performance to the computationally intensive Conformer model. %while utilizes only $13$x fewer computational resources. 

\section{Methodology}

%In this section, we detailed each related proposed component in our customized KWS system.
\label{sec:method}

\subsection{Encoder-Decoder Architecture}

\begin{figure}[t]

\begin{minipage}[t]{1.0\linewidth}
  \centering
  \centerline{\includegraphics[width=8.8cm]{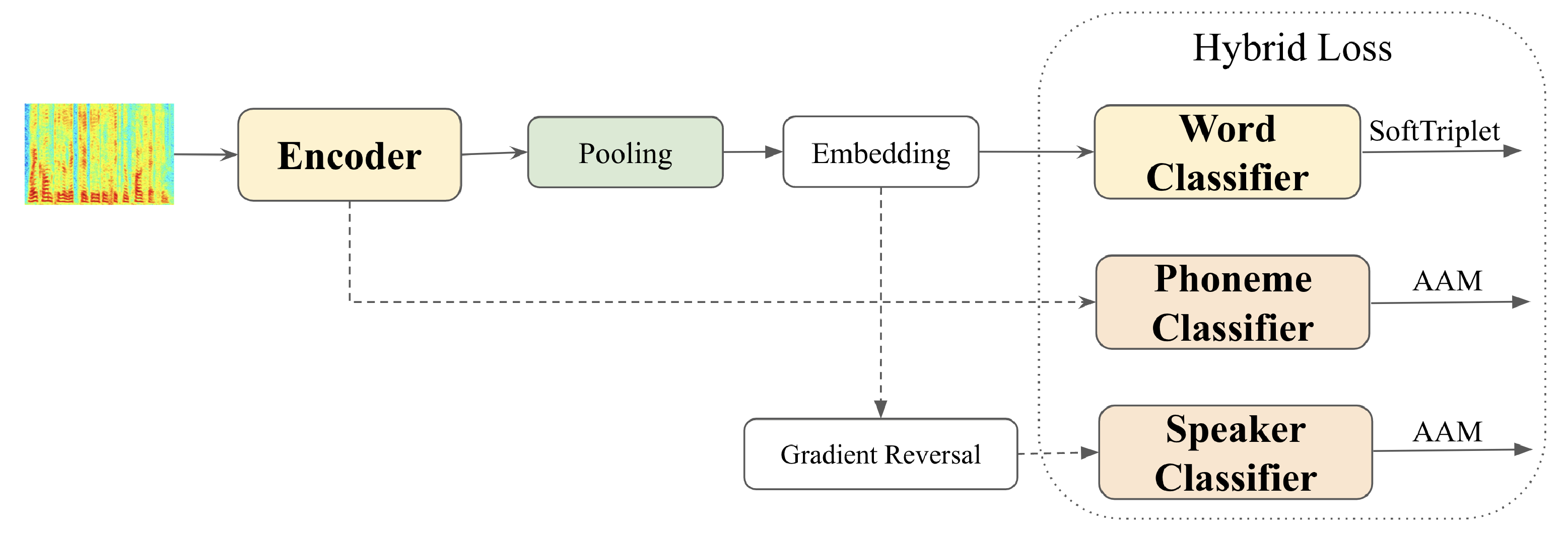}}
%  \vspace{2.0cm}
  % \centerline{(a) Result 1}\medskip
\end{minipage}
 
\caption{The customized KWS training framework.}
\label{fig:new_train}
 
\end{figure}

The system architecture is illustrated in Figure \ref{fig:new_train}. It adopts an encoder-decoder structure during the training phase. The encoder takes the acoustic feature of a word phrase as input and produces an embedding that is subsequently forwarded into the decoder for classification. The pooling layer serves as a dimensionality reduction technique to create a concise yet informative embedding. During the testing phase, a user enrolls in the system by providing a few samples of the customized keyword. Detection occurs by comparing the embedding of the testing speech within a sliding window against the enrolled samples.

\subsection{Feature Encoder}
\label{subsec:encoder}
%We adopt three model architecture options as the feature encoder to map acoustic features to word-level embeddings.
The primary objective of the encoder is to efficiently learn effective embeddings for both acoustic and linguistic information. In this section, we investigate three distinct network architectures for encoder modeling: LiCoNet, which strikes a balance between model effectiveness and efficiency; Conformer, renowned for its exceptional capabilities in sequence modeling; and ECAPA\_TDNN, a widely used backbone for speaker verification (SV).

\subsubsection{Linearized Convolution Network (LiCoNet)} 
LiCoNet represents a hardware-efficient architecture specifically designed for the KWS tasks, as detailed in \cite{yang2022lico}. This architecture is carefully crafted as a streaming convolution network built upon the MobileNetV2 backbone \cite{sandler2018mobilenetv2} for training. It uses equivalent linear operators to ensure efficient inference while preserving a high level of detection accuracy. Each LiCo-Block is structured as a bottleneck configuration composed of three 1D convolution layers. The initial layer employs streaming convolution with a kernel size greater than $1$, followed by two subsequent point-wise convolutions.

\subsubsection{Convolution-augmented Transformer (Conformer)} 
%, \cite{wang2021wake}
The Conformer architecture has proven its remarkable effectiveness within the sequence-to-sequence domain \cite{bello2019attention} and has achieved significant success in the realm of speech recognition tasks \cite{zhang2020transformer} \cite{li2019jasper} \cite{kriman2020quartznet}.  This architecture seamlessly integrates the capabilities of both convolutional and self-attention mechanisms, providing a flexible and exceptionally potent solution for learning feature representations from sequential data.
Each Conformer block comprises four consecutive modules, including a feed-forward module, a self-attention module, a convolution module, and a second feed-forward module \cite{bello2019attention}. This Conformer-based encoder demonstrates the ability to leverage position-specific local features through the convolution module, while simultaneously capturing content-based global interactions through the self-attention module.
%Self-attention-based transformer architecture and convolutions have been successful in Automatic Speech Recognition (ASR) \cite{zhang2020transformer,li2019jasper,kriman2020quartznet} has used a streaming transformer for the keyword detection task. Conformer combines convolution and self-attention, which can be adopted as an effective encoder to model local features and global interactions \cite{bello2019attention}. 
%Each conformer block is comprised of four stacked modules, including a feed-forward module, a self-attention module, a convolution module, and a second feed-forward module \cite{bello2019attention}. The conformer-based encoder is able to exploit position-wise local features (via convolution module) and capture content-based global interactions (via self-attention module). 

\subsubsection{ECAPA\_TDNN} 
The QbyE KWS training and testing process shares similarities with SV. We hence consider ECAPA\_TDNN, a commonly used backbone model architecture for the SV tasks \cite{desplanques2020ecapa}, as a potential choice for encoder modeling in this study. The ECAPA\_TDNN model consists of a 1D convolution followed by three 1D SE-Res2Blocks, a 1D convolution, an attentive statistical pooling, and a fully connected (FC) layer. After each layer within the SE-Res2Block, we apply non-linear ReLU activation and batch normalization (BN). The embedding feature vectors are extracted from the FC layer.

%LiCoNet is hardware-efficient architecture that is tailored to the KWS task \cite{yang2022lico}. It is a carefully-designed streaming convolution network that utilizes equivalent linear operators for efficient inference meanwhile maintains impressive detection accuracy. Each LiCo-Block is a bottleneck structure consisting of three 1D convolution layers: the first layer is a streaming convolution with a kernel size greater than $1$, followed by two point-wise convolutions. 
%To reduce the computational cost, \cite{yang2022lico} proposes the LiCo-Net, which is a hardware-efficient architecture for KWS. The LicoNet is a dual-phase system that uses linear operators at the evaluation stage for efficient inference and employs streaming convolutions at the training phase to maintain a high model capacity \cite{yang2022lico}. 

%A streaming convolution has become a common technique for reducing computational cost, which infers the output at each timestamp based on a small new input from the streaming audio and the previously computed history \cite{alvarez2019end,rybakov2020streaming}. 
\subsection{Feature Aggregator}
Pooling plays an essential role in neural architectures (see Figure \ref{fig:new_train}), serving the purpose of distilling crucial insights from sequential data while preserving essential contextual details. In the context of our study, we investigate two distinct pooling strategies for word embedding learning.
%The output of an encoder is fed into a feature aggregator to map frame-level sequence features to word-level embeddings.

\noindent\textbf{Attentive Statistic Pooling (ASP)} combines the strengths of both statistical pooling and attention mechanisms \cite{okabe2018attentive}. Attention allows the model to dynamically weigh the importance of different elements along the temporal dimension, enabling the extraction of salient features that are crucial for the task at hand.

%uses an attention mechanism to give different weights to each frame of a sequence feature, and then outputs the weighted means and weighted standard deviations \cite{okabe2018attentive}.

\noindent\textbf{Spectral-temporal Graph Attentive Pooling (GAP)} has gained success in the field of speech and audio processing \cite{tak2021end} \cite{tak2021graph}. It leverages the power of graph neural networks to comprehend complex relationships within spectral-temporal data. The spectral and temporal attention module comprises three graph attention blocks, each housing the graph attention network and graph pooling. This configuration empowers the model to adapt the pooling procedure dynamically, facilitating the extraction of crucial features.

%was used for speaker verification anti-spoofing task \cite{tak2021end}. The spectral and temporal attention module is comprised of three graph attention blocks, each of which contains the graph attention network (GAT) and graph pooling. GAT models the relationships residing in different sub-bands or temporal intervals of artifacts \cite{tak2021graph}. Graph pooling selects a subset of the most informative nodes (i.e., discriminative cues in either spectral or temporal domains) to form more discriminative graphs. 

%In the first two parallel blocks (i.e., the spectral and temporal attention blocks), before the GAT layer is applied, temporal and spectral features are collapsed respectively to a single dimension via max-pooling. Then model-level fusion is employed to exploit the complementary information from the previous two graphs. The spectro-temporal attention block operates upon the fused graph to model relationships spanning both domains. 

\subsection{Loss Function}
The learning process of the encoder is designed to maximize the discriminative power of audio embeddings across distinct words \cite{huang2021query} \cite{huang2022qbye}. In this study, we propose multi-task learning that not only considers word-level discrimination but also incorporates fine-grained phoneme information and speaker variability, enhancing the overall modeling effectiveness (see Figure \ref{fig:new_train}).

%\subsubsection{Additive Angular Margin (AAM)}
%\label{sec:aam}

\begin{figure*}[t]
  \begin{minipage}[t]{0.48\textwidth}
  \centering
  \includegraphics[width=7.4cm]{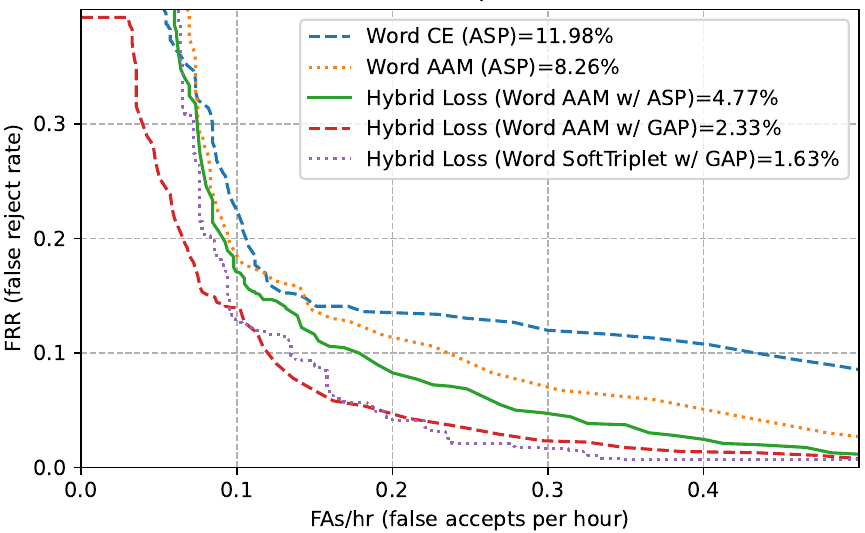}
  \end{minipage}
  \begin{minipage}[t]{0.48\textwidth}
  \centering
  \includegraphics[width=7.4cm]{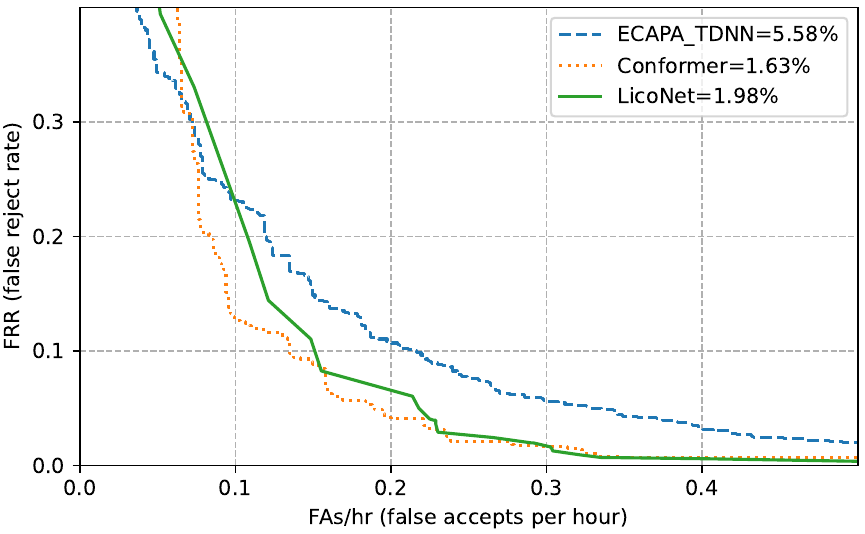}
  \end{minipage}
  \caption{DET curves of Conformer using various loss formulations (left) and of different encoders using the hybrid loss (right).}
  \label{fig:det}
  
\end{figure*}

\subsubsection{Word Discrimination}
For QbyE KWS tasks, it's important that the model possesses strong generalization capabilities. The QbyE system can hence be conceptualized as an optimization problem focusing on minimizing the distance between the embeddings of the same word while simultaneously maximizing the embedding distance of different words. 
We explore two popular losses for this purpose: additive angular margin (AAM) and SoftTriplet.

\noindent\textbf{AAM} emphasizes the angular separation between class embeddings and is prevalent in the context of face recognition and feature embedding \cite{deng2019arcface}. The loss function is defined as,
%has advantages over CE, which optimizes embedding space distribution to enforce compactness for intra-class samples and separation for inter-class samples \cite{deng2019arcface}.
\begin{equation}
\small
\label{eq:aam}
\mathcal{L}_{aam}=-\mathop{\log}\frac{e^{\mathbi{s}\: \cos(\theta_{y_i}+m)}}{e^{\mathbi{s}\: \cos(\theta_{y_i}+m)}+\sum_{j=1,j\ne y_i}^Ce^{\mathbi{s}\: \cos(\theta_{j})}},
\end{equation}
where $\theta_j$ is the angle between the feature $\mathbf{x}_i \in\mathbb{R}^d$ and the weight $\mathbf{w}_j \in\mathbb{R}^d$.  
%($\mathbf{x}_i \in \mathbb{R}^d$, and $d$ is the dimension of word-level embeddings)
$\mathbf{w}_j$ denotes the $j$-th column of the weight $[\mathbf{w}_1, \cdots , \mathbf{w}_C ] \in \mathbb{R}^{d\times C}$ of the last fully-connected layer that maps $d$-dimensional embeddings to the logits. $C$ is the number of classes
%The individual weight $|\mathbf{w}_j|=1$ and feature $|\mathbf{x}_i|$ are fixed by $l_2$ normalization. 
and $\mathbi{s}$ is a rescaling factor. An additive angular margin $m$ is applied for adjustment.

\noindent\textbf{SoftTriplet} has showcased effectiveness for QbyE KWS modeling in \cite{huang2021query}. It combines triplet loss and softmax loss, defined as \cite{qian2019softtriple},
%As introduced in \cite{huang2021query}, we have chosen to adopt the SoftTriplet loss \cite{qian2019softtriple} as a word-level loss function,
%and simultaneously account for intra-class variance—such as speaker variability or variations in speaking rates—we have chosen to adopt the SoftTriplet loss \cite{qian2019softtriple} as a word-level loss function,
%We opt to employ the SoftTriplet loss \cite{qian2019softtriple} as a word-level loss to optimize this objective and meanwhile to model the intra-class variance, e.g., speaker variability or distinct speaking rates,
%To take intra-class variance (e.g., variations due to different speakers or varying durations for the same word), we opt to employ the SoftTriplet loss \cite{qian2019softtriple} as a word-level loss,
%For customized KWS, training and testing samples have no data distribution overlap, which therefore requires the model to have better generalization ability. The QbyE system can be cast as an optimization problem with triplet constraints, where the goal is to minimize the distance between embeddings of a specific enrolled keyword and embeddings of the same word phrase while maximizing the distance corresponding to other word phrases. SoftTriplet loss \cite{qian2019softtriple} considers the real-world situation with intra-class variance (i.e. same word samples collected from different speaker/varied duration), so we adopt SoftTriplet as a word-level loss, which is formulated as,
\begin{equation}
\small
\label{eq:triplet2}
\mathcal{L}_{st}({\mathbf{x}}_i)=-\mathop{\log}\frac{\exp(\lambda(\mathcal{S}^\prime_{i,y_i}-\delta))}{\exp(\lambda(\mathcal{S}^\prime_{i,y_i}-\delta))+\sum_j\exp(\lambda\mathcal{S}^\prime_{i,j})},
\end{equation}
%where $\mathcal{S}^\prime_{i,c}=\sum_k\frac{exp(\frac{1}{\gamma}\mathbf{x}_i^\top\mathbf{w}_c^k)}{\sum_kexp(\frac{1}{\gamma}\mathbf{x}_i^\top\mathbf{w}_c^k)}\mathbf{x}_i^\top\mathbf{w}_c^k$
where $\mathcal{S}^\prime_{i,c}$ is the similarity between feature $\mathbf{x}_i \in \mathbb{R}^d$ and the class $c$. 
%$\mathbf{w}_c^k$ is the $k$-th center out of $K$ centers for the class $c$.
%(revisit Sec. \ref{sec:aam} with the single center for each class). 
$\delta$ is a predefined margin. $\lambda$ denotes a scaling factor.

%The word-level SoftTriplet can be a dominant loss in the hybrid loss to maintain stable classification performance. Performance degradation can also be caused by that keywords being enrolled by different speakers. 
\subsubsection{Speaker Variability} 
Acoustic variations related to individual speakers, such as differences in pitch, tone, or pronunciation, exert a significant influence on speech modeling. Existing approaches in QbyE KWS often assume that the system user is the same as the enrolled speaker. To disentangle speaker dependency within the application, we incorporate a reverse speaker loss into our methodology, with the objective of learning speaker-independent embeddings. Specifically, we design an AAM-based reverse speaker loss, which is employed to maximize the speaker classification loss through the application of a gradient reversal layer (GRL) \cite{ganin2015unsupervised} during the training process. The parameter-free GRL functions as an identity transform during forward propagation but reverses gradients during back-propagation.
%feeding them into the preceding layer. 

\subsubsection{Phoneme Context} 
Phonemes serve as the fundamental phonetic units that compose spoken words. Incorporating the context of phonemes into modeling offers a nuanced source of information for refining word embeddings. In our approach, we introduce a dedicated phoneme classifier into the training framework. Specifically, we adopt the AAM loss for phoneme classification.
The phoneme loss is computed by aggregating the frame-level AAM loss on the phoneme labels across all frames.
%Phonemes represent phonetic elements that constitute spoken words. Incorporating phoneme context into modeling can provide fine-grained information for learning word embedding. As illustrated in Fig.~\ref{fig:new_train}, we introduce a phoneme classifier into the training framework. The phoneme loss is calculated by averaging the frame-level AAM loss of phoneme labels across all the frames.
%Sequence phonetic information can also help to improve system performance by leveraging frame-level acoustic information. 
%We calculate AAM-based loss for each frame with a phoneme label for each, and then average all frame-level loss.

\subsubsection{Multi-Task Learning}
Consequently, the hybrid loss function for multi-task learning is constructed as a combination of word-level loss, the reverse speaker loss, and phoneme-level loss.
%Speaker-dependent acoustic variations such as pitch, tone, or pronunciation, have a profound impact on speech modeling. The existing approaches on QbyE KWS often expects the system user is the same as the enrolled speaker. To disentangle speaker dependency in the application, we employ a reverse speaker loss aiming to learn speaker-independent embedding (see Fig.~\ref{fig:new_train}). Specifically, we formulate an AAM-based reversed speaker loss, which maximizes speaker classification loss  via a gradient reversal layer (GRL) \cite{ganin2015unsupervised} during training.
%To further mitigate this performance degradation, we formulate an AAM-based reversed speaker loss, which maximizes speaker classification loss  via a gradient reversal layer (GRL) \cite{ganin2015unsupervised} during training. 
%The parameter-free GRL acts as an identity transform in the forward propagation. During the back-propagation, GRL reverses the gradients from the subsequent level and feeds into the preceding layer.  
%Therefore, the hybrid loss is formulated as a combination of word-level loss, phoneme-level loss as well as reverse speaker loss,
\begin{equation*}
\small
\begin{split}
\label{eq:hybrid}
 \mathcal{L}(\mathbf{x},\mathbf{y}) = \mathcal{L}\left(\mathbf{x},y^w\right)
 - \eta \space \mathcal{L}_{aam}(\mathbf{x},y^s) + \mu\mathcal{L}_{aam}(\mathbf{x},y^p),
 \end{split}
\end{equation*}
where $\mathbf{x}$ is the acoustic feature vector, and $\mathbf{y} = (y^w, y^s, y^p)$. $y^w$ is the word label, $y^s$ is the speaker label, and $y^p \in \mathbb{R}^T$ is the phoneme label sequence of $T$ frames. $\eta$ and $\mu$ are scaling factors. Note that the word-level loss $\mathcal{L}\left(\mathbf{x},y^w\right)$ can be expressed as either $\mathcal{L}_{aam}$ using AAM or $\mathcal{L}_{st}$ using SoftTriplet.

\section{Experiments}
\label{sec:exp}

\subsection{Dataset}

We use the Librispeech \cite{panayotov2015librispeech} dataset containing 960 hours of read English audiobooks sampled at $16$ kHz along with transcriptions. We employ a pre-trained acoustic model for the force-alignment to segment utterances into individual words. Each word-level segment is standardized to $2$s long by clipping or zero padding on both sides of the audio. We use the internal aggregated and de-identified keyword dataset for evaluation. The positive data contains $275.7$k utterances from $629$ speakers. The total duration of negative data is up to $200$ hours. We extract acoustic features using $40$-dimensional log Mel-filterbank energies computed over a $25$ms window every $10$ms. The evaluation dataset employed in this study is entirely separate from the training data, ensuring that the keywords used for evaluation are not revealed in the training set. This setup aligns with the principles of customized keyword spotting. Additionally, the substantial size of the dataset ensures robust experimental results and conclusive findings.

%237,684 hey Facebook 
%#275,671 utternaces
%#629 speakers
%200.39569722222222hours negative
%50.62424444444444hours positive
\begin{table*}
\fontsize{8}{10}\selectfont
\caption{FRR ($\%$) at $0.3$ FAs/Hr for different loss function formulation, feature pooling strategies and encoder models.}
\vspace{-0.2cm}
%Besides investigating the effectiveness of different encoder models, we employ CE and AAM for word-level loss calculation. Then reversed speaker loss is combined with word-level AAM loss. Additionally, the sequence phoneme is also added to the total hybrid loss. The aforementioned systems are using ASP for feature pooling, we then replace it with GAP. Finally, we replace word-level AAM with SoftTriplet loss. 
    \centering
\begin{tabular}{c||cc||ccc||c}
\toprule[1.5pt]
\multirow{2}{*}{Encoder} & \multicolumn{2}{c||}{Single Loss}                                                                                                        & \multicolumn{3}{c||}{Hybrid Loss (Word AAM)}                                                                                                                                                                                               & Hybrid Loss (Word SoftTriplet)                                    \\ \cline{2-7} 
                         & \multicolumn{1}{c|}{\begin{tabular}[c]{@{}c@{}}Word CE\\ (ASP)\end{tabular}} & \begin{tabular}[c]{@{}c@{}}Word AAM\\ (ASP)\end{tabular} & \multicolumn{1}{c|}{\begin{tabular}[c]{@{}c@{}}Speaker\\ (ASP)\end{tabular}} & \multicolumn{1}{c|}{\begin{tabular}[c]{@{}c@{}}Speaker + Phoneme\\ (ASP)\end{tabular}} & \begin{tabular}[c]{@{}c@{}}Speaker + Phoneme\\ (GAP)\end{tabular} & \begin{tabular}[c]{@{}c@{}}Speaker + Phoneme\\ (GAP)\end{tabular} \\ \toprule[1pt]
ECAPA\_TDNN              & \multicolumn{1}{c|}{$16.28$}                                                   & $12.09$                                                    & \multicolumn{1}{c|}{$10.81$}                                                   & \multicolumn{1}{c|}{$8.95$}                                                              & $7.29$                                                              & $5.58$                                                              \\ 
Conformer                & \multicolumn{1}{c|}{$11.98$}                                                   & $8.26$                                                     & \multicolumn{1}{c|}{$4.88$}                                                    & \multicolumn{1}{c|}{$4.77$}                                                              & $2.33$                                                              & $\mathbf{1.63}$                                                              \\ 
LiCoNet                  & \multicolumn{1}{c|}{$13.20$}                                                   & $9.75$                                                     & \multicolumn{1}{c|}{$7.49$}                                                    & \multicolumn{1}{c|}{$5.36$}                                                              & $3.63$                                                              & $\mathbf{1.98}$                                                              \\ \toprule[1.5pt]
\end{tabular}
    \label{tab:ablation}
\vspace{-0.2cm}
\end{table*}

\subsection{Experimental Setup}
\textbf{Model architecture} We conduct the experiments on three architectures as described in Section \ref{subsec:encoder}: LicoNet, Conformer, ECAPA\_TDNN. We construct LiCoNet by stacking five LiCo-Blocks with the expansion factor of $6$ and the kernel size of $5$ \cite{yang2022lico}. Conformer has two heads per multi-headed self-attention layer with $128$ input and output nodes \cite{shi2021emformer}. The linear hidden units have a dimensionality of $192$, and the convolution module uses the kernel size of $7$. We setup ECAPA\_TDNN with $128$ channels in the convolution layers and a $64$ dimensional bottleneck in the SE-Block and attention module. The scale dimension in the Res2Block is $8$. Table \ref{tab:size} presents the model size and floating point operations per second (FLOPs) of $2$s audio for each encoder model.

\noindent\textbf{Feature aggregator}
We compare GAP against ASP as the feature aggregator. The spectral, temporal and spectro-temporal attention blocks use pooling ratios of $0.71$, $0.86$, and $0.71$.

\noindent\textbf{Loss function} We focus on investigating the effectiveness of different loss formulations. Due to limited space, we only show the best results from SoftTriplet. Similar improvements can also be seen in other setups.
$m$ and $\mathbi{s}$ in Eq. \ref{eq:aam} are set to $0.2$ and $32$. SoftTriplet uses $\lambda=60$, $\delta=0.03$, and $K=10$.
%(when $\lambda$ is large, it dominates the results and the effect of $\gamma$ can be ignored).
The weights $\eta$ and $\mu$ in multi-task learning are set to $0.1$ and $0.5$.

\noindent\textbf{Training and testing protocols} All KWS models are trained to predict 1002 targets (i.e., the top $1$k frequent words, Silence, and Unknown). We use a batch size of 64 with 8 GPUs for 40-epoch training. We adopt the triangular2 policy \cite{smith2017cyclical} using the Adam optimizer with a cyclical learning rate increased from $1$e-$8$ to $1$e-$3$ in $20$k warming-up updates. 
%\cite{kingma2014adam}
During testing, $3$ utterances of any speakers were randomly picked as enrollments. Given a query, the cosine distance is used to compare the similarity between embeddings of the query and the $3$ enrollments. The minimum distance is used to compare against a threshold to make the detection decision. We present the model performance by plotting detection error trade-off (DET) curves, where the x-axis and y-axis represent the number of false accepts (FA) per hour and false reject rate (FRR). 

\subsection{Results and Discussion}
\label{sec:res}
Table \ref{tab:ablation} summarizes FRR for each experiment at $0.3$ FAs/Hr.
In Figure \ref{fig:det}, we present DET curves for Conformer using various loss formulations and pooling strategies (left), and those for different encoders in the optimal multi-task learning setup (right). It is evident that all model architectures achieve their best performance in the hybrid loss configuration using word SoftTriplet and GAP. 
Particularly, LiCoNet demonstrates comparable performance to Conformer in the optimal setup.
%In this study, we investigate different model architectures as the encoder backbone model and compare two feature aggregation strategies, then explore multiple loss function formulations. We showcase the effectiveness of each proposed component with an ablation study and DET curves.

\noindent\textbf{Single Loss vs.~Hybrid Loss} 
In single task learning for word classification, the AAM loss significantly outperforms the CE loss across different encoders. Specifically, AAM improves FRR by $25.7\%$ for ECAPA\_TDNN, $31\%$ for Conformer, and $26.1\%$ for LiCoNet. In the hybrid loss configuration featuring the word AAM loss, the inclusion of the reverse speaker loss greatly decreases FRR, particularly for Conformer, resulting in a reduction of  $40.9\%$. By incorporating the phoneme loss, we can notice additional enhancements. The efficacy of the hybrid loss underscores the value of using complementary information from both speakers and phonemes for QbyE KWS.
%compared to the conventional CE loss, the AAM loss improves the system performance significantly. 
%When the hybrid loss is integrated with the reversed speaker loss, we observe a system boost for all three encoder models, especially the Conformer one. 
%As another component of the hybrid loss, the sequence phoneme loss exploits the complementary acoustic information residing in the input features, which also contributes to the overall system performance and shows consistent benefits for all three encoder models. 

\noindent\textbf{ASP vs.~GAP} In the hybrid loss setup using word AAM loss, GAP delivers further substantial improvements compared to ASP across all encoders. In particular, FRR has been decreased by $18.5\%$ for ECAPA\_TDNN, $51.1\%$ for Conformer, and $32.2\%$ for LiCoNet. 
These improvements align with the enhancement observed in speaker verification \cite{tak2021end} and imply the increased discriminative capability introduced by the graph pooling strategy. 

\noindent\textbf{AAM vs.~SoftTriplet} Further in the hybrid loss configuration employing GAP, the word SoftTriplet loss consistently leads to the best system performance across all models, with a particularly impressive $45.4\%$ reduction in FRR for LiCoNet. This demonstrates the generalizability of SoftTriplet to capture potential unseen intra-variance within the evaluation data. 
%The aforementioned experiments are all using ASP as the pooling method, the KWS system is improved further with a lower false alarm rate. 
%Eventually, we make a substitution for the word-level AAM with the SoftTriplet loss to capture potential unseen intra-variance of evaluation data, which achieves the best system performance across all three encoder models. 

\noindent\textbf{Encoder Effectiveness} Conformer consistently maintains superior performance across various loss formulations and pooling strategies. However, despite the inherent capacity limitations of linear operators in LiCoNet when compared to the attention scheme in Conformer, LiCoNet achieves comparable performance to Conformer ($1.98\%$ vs.~$1.63\%$ FRR) in the best multi-task learning setup that employs the word SoftTriplet loss and GAP. 
This observation demonstrates the effectiveness of the proposed QbyE framework in maximizing the potential of simpler models, making them capable of delivering results on par with their more complex counterparts. 

%For model-wise comparison, the ECAPA\_TDNN offers a decent system performance in terms of false rejection rate, while the Conformer outperforms the other two encoder models. Moreover, compared to the Conformer, the LicoNet merely shows a marginal system performance degradation. 
%In Fig. \ref{fig:det} (a), we present DET curves of different loss function formulations and pooling strategies while using Conformer as the feature encoder. In Fig. \ref{fig:det} (b), we present DET curves of systems equipped with different feature encoders while using GAP as the pooling method and the hybrid loss (see in Eq. \ref{eq:hybrid}).
\begin{table}[h]
\fontsize{8}{10}\selectfont
\caption{Encoder model size and computation on a $2$s audio.}
\vspace{-0.2cm}
    \centering
    \begin{tabular}{c|cc}
 % \hline
 \toprule[1.5pt]
 Encoder & $\#$Params & FLOPs \\
 \toprule[1pt]
     %ECAPA\_TDNN & $540.1$K &$39064192$\\
     %LicoNet  & $694.1$K  & $46507824$\\
     %Conformer & $1.37$M & $642176000$ \\
     ECAPA\_TDNN & $540$K &$39.1$M\\
     Conformer & $1.4$M & $642.2$M \\
     LicoNet  & $694$K  & $46.5$M\\
 %& $632.9$K    
\toprule[1.5pt]
\end{tabular}
    \label{tab:size}
\vspace{-0.2cm}
\end{table}

\noindent\textbf{Model Efficiency}  
%Since most on-device KWS system is always-on, it is highly desirable to make the running model hardware-efficient to save memory space and to prolong battery life, while maintaining effective system performance. 
To assess memory and computation efficiency, we present the model size and computational cost (FLOPs) in Table \ref{tab:size}. It is evident that ECAPA\_TDNN exhibits the highest efficiency but has limited performance. 
%The conformer model has the superior ability to model the keyword discriminability, while the high computational cost is not prohibitive. 
On the other hand, Conformer boasts the largest model size with considerably more computational demands, despite its superiority on high model capacity. LiCoNet strikes a favorable balance between model efficiency and effectiveness. It achieves performance on par with Conformer while maintaining computational costs similar to ECAPA\_TDNN. Note that the Conformer model size is considerably larger than the other two models. This results from a trade-off between model size and performance. We observe a significant performance degradation when reducing the size of Conformer, whereas only slight performance improvements are observed when increasing the size of ECAPA\_TDNN and LiCoNet.

% Considering the trade-off between efficiency and efficacy, each model size is constrained to their own minimum with an effective system performance, resulting in the optimal setup for each model. Specifically, we found that there was no significant performance gain while doubling the model size (i.e., increasing convolutional kernel size) of the ECAPA\_TDNN/LiCoNet. In contrast, we saw a significant performance degradation while shrinking the Conformer model size (i.e., decreasing the number of self-attention layers/the hidden unit size).

\section{Conclusion}
\label{sec:cons}
In this study, we introduce a novel QbyE KWS system that employs a spectral-temporal graph pooling layer and multi-task learning. This framework aims to effectively learn speaker-invariant and linguistic-informative embeddings for QbyE KWS tasks. Within this framework, we investigate three distinct network architectures for encoder modeling: LiCoNet, Conformer and ECAPA\_TDNN. The experimental results showcase the effectiveness of the proposed QbyE framework in maximizing the potential of simpler models such as LiCoNet, making them capable of delivering results on par with their more complex counterparts.

%The experimental results show that the hardware-efficient LiCoNet performs at a similar level to the computationally expensive Conformer, and the proposed hybrid loss function effectively enhances model effectiveness.  
%We investigated the effectiveness of three different model architectures as the feature encoder, different pooling strategies to form the compact but informative embedding. Furthermore, we explored various loss function formulations to exploit discriminative information residing in the acoustic features. 
%Experimental results have demonstrated that the LicoNet shows a better trade-off between hardware efficiency and model accuracy; spectro-temporal GAP serves as an effective feature aggregator; the proposed hybrid loss acts as the best-performing option for customized KWS modeling.

\bibliographystyle{IEEEtran}
\bibliography{template}

% Generated by IEEEtran.bst, version: 1.13 (2008/09/30)
\begin{thebibliography}{10}
\providecommand{\url}[1]{#1}
\csname url@samestyle\endcsname
\providecommand{\newblock}{\relax}
\providecommand{\bibinfo}[2]{#2}
\providecommand{\BIBentrySTDinterwordspacing}{\spaceskip=0pt\relax}
\providecommand{\BIBentryALTinterwordstretchfactor}{4}
\providecommand{\BIBentryALTinterwordspacing}{\spaceskip=\fontdimen2\font plus
\BIBentryALTinterwordstretchfactor\fontdimen3\font minus
  \fontdimen4\font\relax}
\providecommand{\BIBforeignlanguage}[2]{{%
\expandafter\ifx\csname l@#1\endcsname\relax
\typeout{** WARNING: IEEEtran.bst: No hyphenation pattern has been}%
\typeout{** loaded for the language `#1'. Using the pattern for}%
\typeout{** the default language instead.}%
\else
\language=\csname l@#1\endcsname
\fi
#2}}
\providecommand{\BIBdecl}{\relax}
\BIBdecl

\bibitem{chen2015query}
G.~Chen, C.~Parada, and T.~N. Sainath, ``Query-by-example keyword spotting
  using long short-term memory networks,'' in \emph{ICASSP}.\hskip 1em plus
  0.5em minus 0.4em\relax IEEE, 2015, pp. 5236--5240.

\bibitem{chen2018compact}
M.~Chen, S.~Zhang, M.~Lei, Y.~Liu, H.~Yao, and J.~Gao, ``Compact feedforward
  sequential memory networks for small-footprint keyword spotting.'' in
  \emph{Interspeech}, 2018, pp. 2663--2667.

\bibitem{guo2018time}
J.~Guo, K.~Kumatani, M.~Sun, M.~Wu, A.~Raju, N.~Str{\"o}m, and A.~Mandal,
  ``Time-delayed bottleneck highway networks using a dft feature for keyword
  spotting,'' in \emph{ICASSP}.\hskip 1em plus 0.5em minus 0.4em\relax IEEE,
  2018, pp. 5489--5493.

\bibitem{shan2018attention}
C.~Shan, J.~Zhang, Y.~Wang, and L.~Xie, ``Attention-based end-to-end models for
  small-footprint keyword spotting,'' \emph{arXiv preprint arXiv:1803.10916},
  2018.

\bibitem{wang2019adversarial}
X.~Wang, S.~Sun, C.~Shan, J.~Hou, L.~Xie, S.~Li, and X.~Lei, ``Adversarial
  examples for improving end-to-end attention-based small-footprint keyword
  spotting,'' in \emph{ICASSP}.\hskip 1em plus 0.5em minus 0.4em\relax IEEE,
  2019, pp. 6366--6370.

\bibitem{hazen2009query}
T.~J. Hazen, W.~Shen, and C.~White, ``Query-by-example spoken term detection
  using phonetic posteriorgram templates,'' in \emph{ASRU}.\hskip 1em plus
  0.5em minus 0.4em\relax IEEE, 2009, pp. 421--426.

\bibitem{zhang2009unsupervised}
Y.~Zhang and J.~R. Glass, ``Unsupervised spoken keyword spotting via segmental
  dtw on gaussian posteriorgrams,'' in \emph{ASRU}.\hskip 1em plus 0.5em minus
  0.4em\relax IEEE, 2009, pp. 398--403.

\bibitem{anguera2013memory}
X.~Anguera and M.~Ferrarons, ``Memory efficient subsequence dtw for
  query-by-example spoken term detection,'' in \emph{ICME}.\hskip 1em plus
  0.5em minus 0.4em\relax IEEE, 2013, pp. 1--6.

\bibitem{bluche2020small}
T.~Bluche, M.~Primet, and T.~Gisselbrecht, ``Small-footprint open-vocabulary
  keyword spotting with quantized lstm networks,'' \emph{arXiv preprint
  arXiv:2002.10851}, 2020.

\bibitem{zhuang2016unrestricted}
Y.~Zhuang, X.~Chang, Y.~Qian, and K.~Yu, ``Unrestricted vocabulary keyword
  spotting using lstm-ctc.'' in \emph{Interspeech}, 2016, pp. 938--942.

\bibitem{huang2021query}
J.~Huang, W.~Gharbieh, H.~S. Shim, and E.~Kim, ``Query-by-example keyword
  spotting system using multi-head attention and soft-triple loss,'' in
  \emph{ICASSP}.\hskip 1em plus 0.5em minus 0.4em\relax IEEE, 2021, pp.
  6858--6862.

\bibitem{huang2022qbye}
J.~Huang, W.~Gharbieh, Q.~Wan, H.~S. Shim, and C.~Lee, ``Qbye-mlpmixer:
  Query-by-example open-vocabulary keyword spotting using mlpmixer,''
  \emph{arXiv preprint arXiv:2206.13231}, 2022.

\bibitem{tak2021end}
H.~Tak, J.-w. Jung, J.~Patino, M.~Kamble, M.~Todisco, and N.~Evans,
  ``End-to-end spectro-temporal graph attention networks for speaker
  verification anti-spoofing and speech deepfake detection,'' \emph{arXiv
  preprint arXiv:2107.12710}, 2021.

\bibitem{xiang2019margin}
X.~Xiang, S.~Wang, H.~Huang, Y.~Qian, and K.~Yu, ``Margin matters: Towards more
  discriminative deep neural network embeddings for speaker recognition,'' in
  \emph{2019 Asia-Pacific Signal and Information Processing Association Annual
  Summit and Conference (APSIPA ASC)}.\hskip 1em plus 0.5em minus 0.4em\relax
  IEEE, 2019, pp. 1652--1656.

\bibitem{deng2019arcface}
J.~Deng, J.~Guo, N.~Xue, and S.~Zafeiriou, ``Arcface: Additive angular margin
  loss for deep face recognition,'' in \emph{CVPR}, 2019, pp. 4690--4699.

\bibitem{yang2022lico}
H.~Yang, Z.~Yang, L.~Wan, B.~Zhang, Y.~Shi, Y.~Huang, I.~Enchev, L.~Tang,
  R.~Alvarez, M.~Sun \emph{et~al.}, ``Lico-net: Linearized convolution network
  for hardware-efficient keyword spotting,'' \emph{arXiv preprint
  arXiv:2211.04635}, 2022.

\bibitem{sandler2018mobilenetv2}
M.~Sandler, A.~Howard, M.~Zhu, A.~Zhmoginov, and L.-C. Chen, ``Mobilenetv2:
  Inverted residuals and linear bottlenecks,'' in \emph{Proceedings of the IEEE
  conference on computer vision and pattern recognition}, 2018, pp. 4510--4520.

\bibitem{bello2019attention}
I.~Bello, B.~Zoph, A.~Vaswani, J.~Shlens, and Q.~V. Le, ``Attention augmented
  convolutional networks,'' in \emph{ICCV}, 2019, pp. 3286--3295.

\bibitem{zhang2020transformer}
Q.~Zhang, H.~Lu, H.~Sak, A.~Tripathi, E.~McDermott, S.~Koo, and S.~Kumar,
  ``Transformer transducer: A streamable speech recognition model with
  transformer encoders and rnn-t loss,'' in \emph{ICASSP}.\hskip 1em plus 0.5em
  minus 0.4em\relax IEEE, 2020, pp. 7829--7833.

\bibitem{li2019jasper}
J.~Li, V.~Lavrukhin, B.~Ginsburg, R.~Leary, O.~Kuchaiev, J.~M. Cohen,
  H.~Nguyen, and R.~T. Gadde, ``Jasper: An end-to-end convolutional neural
  acoustic model,'' \emph{arXiv preprint arXiv:1904.03288}, 2019.

\bibitem{kriman2020quartznet}
S.~Kriman, S.~Beliaev, B.~Ginsburg, J.~Huang, O.~Kuchaiev, V.~Lavrukhin,
  R.~Leary, J.~Li, and Y.~Zhang, ``Quartznet: Deep automatic speech recognition
  with 1d time-channel separable convolutions,'' in \emph{ICASSP}.\hskip 1em
  plus 0.5em minus 0.4em\relax IEEE, 2020, pp. 6124--6128.

\bibitem{desplanques2020ecapa}
B.~Desplanques, J.~Thienpondt, and K.~Demuynck, ``Ecapa-tdnn: Emphasized
  channel attention, propagation and aggregation in tdnn based speaker
  verification,'' \emph{arXiv preprint arXiv:2005.07143}, 2020.

\bibitem{okabe2018attentive}
K.~Okabe, T.~Koshinaka, and K.~Shinoda, ``Attentive statistics pooling for deep
  speaker embedding,'' \emph{arXiv preprint arXiv:1803.10963}, 2018.

\bibitem{tak2021graph}
H.~Tak, J.-w. Jung, J.~Patino, M.~Todisco, and N.~Evans, ``Graph attention
  networks for anti-spoofing,'' \emph{arXiv preprint arXiv:2104.03654}, 2021.

\bibitem{qian2019softtriple}
Q.~Qian, L.~Shang, B.~Sun, J.~Hu, H.~Li, and R.~Jin, ``Softtriple loss: Deep
  metric learning without triplet sampling,'' in \emph{Proceedings of the
  IEEE/CVF International Conference on Computer Vision}, 2019, pp. 6450--6458.

\bibitem{ganin2015unsupervised}
Y.~Ganin and V.~Lempitsky, ``Unsupervised domain adaptation by
  backpropagation,'' in \emph{International conference on machine
  learning}.\hskip 1em plus 0.5em minus 0.4em\relax PMLR, 2015, pp. 1180--1189.

\bibitem{panayotov2015librispeech}
V.~Panayotov, G.~Chen, D.~Povey, and S.~Khudanpur, ``Librispeech: an asr corpus
  based on public domain audio books,'' in \emph{ICASSP}.\hskip 1em plus 0.5em
  minus 0.4em\relax IEEE, 2015, pp. 5206--5210.

\bibitem{shi2021emformer}
Y.~Shi, Y.~Wang, C.~Wu, C.-F. Yeh, J.~Chan, F.~Zhang, D.~Le, and M.~Seltzer,
  ``Emformer: Efficient memory transformer based acoustic model for low latency
  streaming speech recognition,'' in \emph{ICASSP}.\hskip 1em plus 0.5em minus
  0.4em\relax IEEE, 2021, pp. 6783--6787.

\bibitem{smith2017cyclical}
L.~N. Smith, ``Cyclical learning rates for training neural networks,'' in
  \emph{WACV}.\hskip 1em plus 0.5em minus 0.4em\relax IEEE, 2017, pp. 464--472.

\end{thebibliography}

\end{document}